\RequirePackage{fix-cm}
\documentclass[smallextended]{svjour3}       % onecolumn (second format)
\smartqed  % flush right qed marks, e.g. at end of proof
\usepackage{graphicx}
\usepackage{amsfonts}
\usepackage{mathrsfs}
\usepackage{amssymb}
\usepackage{bbm}
\usepackage{algorithm}
\usepackage{algorithmic}
\usepackage{listings}
\usepackage{longtable}
\usepackage{pdflscape}
\usepackage{supertabular}
\usepackage{setspace}
\usepackage{verbatim}
\usepackage{wrapfig}
\usepackage{array}
\usepackage{amsmath}
\usepackage{url}
\usepackage[final]{pdfpages}
\usepackage{times}
\usepackage{textcomp}
\usepackage{soul}
\usepackage{booktabs}
\begin{document}

\title{Enhash: A Fast Streaming Algorithm For Concept Drift Detection}
%\subtitle{Do you have a subtitle?\\ If so, write it here}

%\titlerunning{Short form of title}        % if too long for running head

\author{Aashi Jindal \and
        Prashant Gupta \and
        Debarka Sengupta \and
        Jayadeva}

%\authorrunning{Short form of author list} % if too long for running head

\institute{A. Jindal \at
              Department of Electrical Engineering, Indian Institute of Technology Delhi, Hauz Khas, Delhi 110016, India \\
              %Tel.: +123-45-678910\\
              %Fax: +123-45-678910\\
              \email{aashi.jindal@ee.iitd.ac.in}           %  \\
%             \emph{Present address:} of F. Author  %  if needed
           \and
           P. Gupta \at
              Department of Electrical Engineering, Indian Institute of Technology Delhi, Hauz Khas, Delhi 110016, India\\
              \email{prashant.gupta@ee.iitd.ac.in}
            \and
            D. Sengupta \at
            Department of Computer Science and Engineering, Indraprastha Institute of Information Technology, Delhi 110020, India\\
            Infosys Center for Artificial Intelligence, Indraprastha Institute of Information Technology, Delhi 110020, India\\
            Center for Computational Biology, Indraprastha Institute of Information Technology, Delhi 110020, India\\
            \email{debarka@iiitd.ac.in}
            \and
            Jayadeva \at
            Department of Electrical Engineering, Indian Institute of Technology Delhi, Hauz Khas, Delhi 110016, India\\
            \email{jayadeva@ee.iitd.ac.in}
            }

\date{Received: date / Accepted: date}
% The correct dates will be entered by the editor

\newcolumntype{L}[1]{>{\raggedright\let\newline\\\arraybackslash\hspace{0pt}}m{#1}}
\newcolumntype{C}[1]{>{\centering\let\newline\\\arraybackslash\hspace{0pt}}m{#1}}
\newcolumntype{R}[1]{>{\raggedleft\let\newline\\\arraybackslash\hspace{0pt}}m{#1}}

\maketitle

\begin{abstract}
We propose Enhash, a fast ensemble learner that detects \textit{concept drift} in a data stream. A stream may consist of abrupt, gradual, virtual, or recurring events, or a mixture of various types of drift. Enhash employs projection hash to insert an incoming sample. We show empirically that the proposed method has competitive performance to existing ensemble learners in much lesser time. Also, Enhash has moderate resource requirements. Experiments relevant to performance comparison were performed on 6 artificial and 4 real data sets consisting of various types of drifts.
%in performance and speed. All experiments related to performance comparison were performed on 6 artificial and 4 real data sets consisting of various types of drifts.
\end{abstract}

\section{Introduction}
A data stream environment is often characterized by large volumes of data that flow rapidly and continuously. They are processed in an \textit{online} fashion to accommodate data that cannot reside in main memory. A streaming data environment is commonly used for tasks such as making recommendations for users on streaming platforms~\cite{subbian2016recommendations}, and real-time analysis inside IoT devices~\cite{atzori2010internet}. In such a stream, the underlying data distribution may change, and this phenomenon is referred to as \textit{concept drift}. Formally, the posterior probability of a sample's class changes with time. Consequently, the method must also be able to adapt to the new distribution. To adapt to a new \textit{concept}, the method may require supplemental or replacement learning. Tuning a model with new information is termed as supplemental learning. Replacement learning refers to the case when the model's old information becomes irrelevant, and is replaced by new information. A shift in the likelihood of observing a data point $x$ within a particular class when class boundaries are altered, is called \textit{real concept drift}. \textit{Concept drift} without an overlap of true class boundaries, or an incomplete representation of the actual environment, is referred to as \textit{virtual concept drift}. In \textit{virtual concept drift}, one requires supplemental learning, while \textit{real concept drift} requires replacement learning~\cite{gama2014survey}. The other common way to categorize \textit{concept drift} is determined by the speed with which changes occur~\cite{de2019overview}. Hence, drift may be \textit{incremental}, \textit{abrupt} or \textit{gradual}. A \textit{reoccurring drift} is one that emerges repeatedly. Thus, in order to handle \textit{concept drift}, a model must be adaptive to non-stationary environments.

Several methods have been recently proposed to handle \textit{concept drift} in a streaming environment. The most popular of these are ensemble learners~\cite{learn++,learnNse,onlineSmotebagging,leverageBagging,arf,accuracyWeightedEnsemble,dwm,additiveExpertEnsemble}. As the data stream evolves, an ensemble method selectively retains a few learners to maintain prior knowledge while discarding and adding new learners to learn new knowledge. Thus, an ensemble method is quite flexible, and maintains the \textit{stability-plasticity} balance~\cite{lim2003online} i.e. retaining the previous knowledge (\textit{stability}) and learning new concepts (\textit{plasticity}). %Ensemble methods, in practice, offer the best bias-variance trade-offs~\cite{breiman1996arcing,bauer1999empirical}.

In this paper, we propose \textit{Enhash}, an ensemble learner that employs projection hash~\cite{indyk1998approximate} to handle \textit{concept drift}. For incoming samples, it generates a hash code such that similar samples tend to hash into the same bucket. A gradual forgetting factor weights the contents of a bucket. Thus, the contents of a bucket are relevant as long as the incoming stream belongs to the \textit{concept} represented by them. %In case a new concept arrives, then the samples will most likely hash to a different bucket. %For samples with a different concept, a different hash code will be generated.

The rest of the paper is organized as follows. Section~\ref{section} discusses the basic principle of some relevant ensemble learners. Section~\ref{method} elaborates the proposed ensemble learner. Section~\ref{setup} highlights the performance metrics used to evaluate the performances of several ensemble learners on artificial and real data sets with varying \textit{concept drift}. Section~\ref{sec::tuningEnhash} discusses the impact of tuning of parameters on the performance of Enhash. Section~\ref{results} compares the performances of discussed ensemble learners. Section~\ref{sec::ablation} compares the performance of Enhash with its two different variants and highlights the importance of design choice. Section~\ref{conclusion} concludes the aspects of the proposed method.

\section{Related Work}\label{section}
In this section, we discuss some of the widely used ensemble learners for drift detection.

\textit{Learn++}~\cite{learn++} is an \textit{Adaboost}~\cite{freund1997decision,schapire1998boosting,schapire1990strength} inspired algorithm, that generates an ensemble of weak classifiers, each trained on a different distribution of training samples. The multiple classifier outputs are combined using weighted majority voting. For incremental learning, Learn++ updates the distribution for subsequent classifiers such that instances from new classes are given more weights.

While Learn++ is suitable for incremental learning, $\textit{Learn}^{++}\textit{.NSE}$~\cite{learnNse} employs a passive drift detection mechanism to handle non-stationary environments. If the data from a new \textit{concept} arrives and is misclassified by the existing set of classifiers, then a new classifier is added to handle this misclassification. $\textit{Learn}^{++}\textit{.NSE}$ is an ensemble-based algorithm that uses weighted majority voting, where the weights are dynamically adjusted based on classifiers' performance.

\textit{Accuracy-Weighted Ensemble}~\cite{accuracyWeightedEnsemble} is an ensemble of weighted classifiers where the weight of a classifier is inversely proportional to its expected error.

\textit{Additive Expert Ensemble}~\cite{additiveExpertEnsemble} uses a weighted vote of experts to handle \textit{concept drift}. The weights of the experts that misclassify a sample are decreased by a multiplicative constant $\beta \in [0,1]$. In case the overall prediction is incorrect, new experts are added as well.

\textit{Dynamic Weighted Majority (DWM)}~\cite{dwm} dynamically adds and removes classifiers to cope with \textit{concept drift}.

Online bagging and boosting~\cite{oza2005online} ensemble classifiers are used in combination with different algorithms such as \textit{ADaptive WINdowing (ADWIN)}~\cite{adwin} for \textit{concept drift} detection. ADWIN dynamically updates the window by growing it when there is no apparent change, and shrinking it when the data evolves. In general, bagging is more robust to noise than boosting~\cite{dietterich2000experimental,oza2001experimental}. Henceforth, we discuss methods based on online bagging, that approximates batch-bagging by training every base model with $K$ copies of a training sample, where $K \sim \textit{Poisson(1)}$. We refer to the combination of online bagging classifiers with ADWIN as \textit{Online Bagging-ADWIN}.

\textit{Leveraging bagging}~\cite{leverageBagging} exploits the performance of bagging by increasing randomization. Resampling with replacement is employed in online bagging using \textit{Poisson(1)}. Leveraging bagging increases the weights of this resampling by using a larger value of $\lambda$ to compute the \textit{Poisson} distribution. It also increases randomization at the output by using output codes.

\textit{Online SMOTE Bagging}~\cite{onlineSmotebagging} oversamples a minor class by using SMOTE~\cite{chawla2002smote} at each bagging iteration to handle class imbalance. SMOTE generates synthetic examples by interpolating minor class examples.

\textit{Adaptive Random Forest, ARF}~\cite{arf} is an adaptation of Random Forest~\cite{breiman2001random} for evolving data streams. It trains a background tree when there is a warning, and replaces the primary model if drift occurs.

\section{The Proposed Method}\label{method}

Several recent methods employ hashing for online learning and outlier detection~\cite{sathe2016subspace,jindal2018discovery}. We propose Enhash, an ensemble learner, that employs hashing for \textit{concept drift} detection. Let $x_{t}\in \mathbb{R}^{d}$ represent a sample from a data stream $S$ at time step $t$ and let $y\in \{1,2,...,C\}$ represent its corresponding \textit{concept}, where $C$ is the total number of \textit{concepts}. Further, let us assume a family of hash functions $H$ such that $\forall h_{l}\in H$, it maps $x_{t}$ to an integer value. The hash code $h_{l}(x_{t})$ is assigned to $x_{t}$ by hash function $h_{l}$. A bucket is a set of samples with the same hash code; both these terms are used interchangeably. The total number of samples in bucket $h_{l}(x_t)$ is denoted by $N_{h_{l}(x_t)}$. Further assume that $N$ samples have been seen and hashed from the stream so far. Amongst $N$, $N_{c}$ samples belong to the \textit{concept class} $c$ such that $\sum_{c=1}^{c=C}N_{c} = N$. Based on the evidence from the data stream seen so far, the probability of bucket $h_{l}(x_{t+1})$ is given by
\begin{align}
\label{eq::probH}
p(h_{l}(x_{t+1})) = \frac{N_{h_{l}(x_{t+1})}}{N}
\end{align}
and prior for class $c$ is given by
$p(c) = N_{c}/N$. Assuming, $(N_{h_{l}(x_{t+1})})_{c}$ represents the samples of \textit{concept} $c$ in bucket $h_{l}(x_{t+1})$. Hence, the likelihood of $x_{t+1}$ belonging to \textit{concept} $c$ in bucket $h_{l}(x_{t+1})$ is given by
\begin{align}
\label{eq::likelihood}
p(h_{l}(x_{t+1}) | c) = \frac{(N_{h_{l}(x_{t+1})})_c}{N_{c}}
\end{align}
The probability of $x_{t+1}$ belonging to class $c$ is given by
\begin{align}
\label{eq::posterior}
p(c|h_{l}(x_{t+1})) = \frac{p(h_{l}(x_{t+1}) | c)p(c)}{p(h_{l}(x_{t+1}))} = \frac{(N_{h_{l}(x_{t+1})})_{c}}{N_{h_{l}(x_{t+1})}}
\end{align}
Equation~(\ref{eq::posterior}) is simply the normalization of counts in bucket $h_{l}(x_{t+1})$.

To predict the \textit{concept class} of $x_{t+1}$, an ensemble of $L$ such hash functions can be employed and the weight for each \textit{concept class} is computed as
\begin{align}
\label{eq::classweight}
\hat{p}_{c} = \sum_{l=1}^{l=L} \log\Big(1 + p(c|h_{l}(x_{t+1}))\Big)
\end{align}
and the \textit{concept class} is predicted as
\begin{align}
\label{eq::class}
\hat{y} = \arg\max_{c\in\{1,..,C\}}\hat{p}_{c}
\end{align}
To accommodate an incoming sample of class $c$, the bucket is updated as
\begin{align}
\label{eq::update}
(N_{h_{i}(x_{t+1})})_{c} = 1 +  (N_{h_{i}(x_{t+1})})_{c}\:\forall c\in\{1,..,C\}
\end{align}

Enhash utilizes a simple strategy as described above to build an ensemble learner for \textit{concept drift} detection.

\begin{figure}
    \centering
    \includegraphics[width=\textwidth,keepaspectratio]{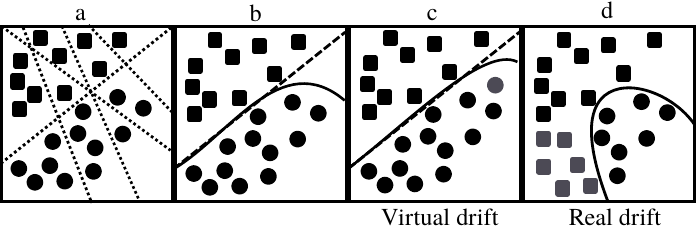}
    \caption{Enhash accommodates both virtual and real drift.}
    \label{fig:drift}
\end{figure}

In Enhash, the projection hash family is selected as a base learner. Here, a hash function involves the dot product of hyperplane $w^{(l)}$ and sample $x_{t}$. It is defined as
\begin{align}
\label{eq::hash}
h_{l}(x_{t}) = \lfloor\frac{1}{\textit{bin-width}}\Bigg(\sum_{j=1}^{j=d}\big(w^{(l)}_{j}*(x_{t})_{j}\big) + \text{bias}^{(l)}\Bigg)\rfloor
\end{align}
where, \textit{bin-width} is quantization width, $w^{(l)}_{j}$ and bias$^{(l)}$ can be sampled from any desired distribution. In our implementation, $w^{(l)}_{j}\sim N(0,1)$ and \\$\text{bias}^{(l)}\sim[-\textit{bin-width}, \textit{bin-width}]$.

Effectively, each $h_{l}$ (\ref{eq::hash})  divides the space into equally spaced unbounded regions of size \textit{bin-width} (earlier referred to as bucket). Equation~(\ref{eq::posterior}) computes the probability of each \textit{concept class} in a region. An ensemble of hash functions makes all the regions bounded. The weight of a \textit{concept class} in the bounded region is computed using (\ref{eq::classweight}). An absolute value of \textit{concept class} is assigned to each region in (\ref{eq::class}). Figure~\ref{fig:drift}a shows the arrangement of randomly generated hyperplanes. The solid line in Figure~\ref{fig:drift}b shows the inferred decision boundary (learned distribution) after an absolute assignment of \textit{concept class} to every bounded region. The dashed line in Figure~\ref{fig:drift}b depicts the true decision boundary (true distribution).

Assuming at time step $t$, Figure~\ref{fig:drift}b shows the current stage of learner. At step $t+1$, a new sample $x_{t+1}$ arrives (gray sample in Figure~\ref{fig:drift}c). After updating the bucket (\ref{eq::update}), the learned distribution shifts and moves towards the true distribution. Hence, Enhash accommodates virtual drift present in the data stream.

Figure~\ref{fig:drift}d depicts the real drift when the true distribution evolves. This requires forgetting some of the previously learned information. Suppose that sample $x_{t+\Delta t}$ hashes to bucket $h_{l}(x_{t+\Delta t})$ at time say, $t + \Delta t$. Assume that previously, $x_t$ from a different \textit{concept}, was hashed to this bucket. In order to accommodate forgetting, Enhash employs a decay factor multiplier to $p(c|h_{l}(x_{t+\Delta t}))$ (\ref{eq::classweight}) while weighting a bounded region.
\begin{align}
\label{eq::updatePosterior}
\hat{p}_c = \sum_{l=1}^{l=L}\log\Big(1 + 2^{-\lambda \Delta t} p(c|h_{l}(x_{t+\Delta t}))\Big)
\end{align}
where $\lambda$ is the decay rate.
The update rule for the bucket (\ref{eq::class}) is also changed to reflect the new \textit{concept class} as follows
\begin{align}
\label{eq::updateBucketNew}
(N_{h_{l}(x_{t+1})})_c = 1+ 2^{-\lambda \Delta t} (N_{h_{l}(x_{t+1})})_c
\end{align}
Setting $\lambda=0$ will reduce these equations to the base case.

In order to break ties in $p(c|h_{l}(x_{t}))$, the class weight in the region is also weighted by the distance of the sample $x_{t}$ from the mean of class samples in bucket $h_{l}(x_{t})$, i.e. $\text{mean}_{h_{l}(x_t)}^{c}$
\begin{align}
\label{eq::distance}
\text{dist}_{c}(x_{t}) = \sqrt{\sum_{i=1}^{i=d}\Big((x_{t})_{i} - (\text{mean}_{h_{l}(x_t)}^{c})_{i}\Big)^{2}}
\end{align}
\begin{align}
\label{eq::finalUpdate}
\hat{p}_c = \sum_{l=1}^{l=L}\log\Big(1 + \frac{2^{-\lambda \Delta t} p(c|h_{l}(x_{t}))}{dist_{c}(x_{t})})\Big)
\end{align}

and remaining ties are broken arbitrarily.

\subsection{Implementation Details}
Algorithm~\ref{algo::methodEdit} gives Enhash's pseudo-code. We refer to an instance of the ensemble as an estimator. The parameter \textit{bin-width} determines the granularity of the buckets, and $\lambda$ represents the rate of decay that accounts for gradual forgetting~\cite{klinkenberg2004learning}. %An estimator is associated with a weight vector $w^{(l)}$ and a bias $bias^{(l)}$.
Every estimator $l$ generates a hash code for incoming stream samples and hashes them into bucket $b^{(l)}$. The hash code for a sample $x$ is compute using (\ref{eq::hash}).
%The projection of the sample $x$ on $w^{(l)}$, shifted by $bias^{(l)}$, and scaled by \textit{bin-width}, gives the hash code $h^{(l)}(x)$.
The timestamp when a sample of class $y$ was last hashed into a bucket $b^{(l)}$ is stored in $tstamp^{(l)}[b^{(l)}][y]$.  $tstamp^{(l)}$ is an infinitely indexable 2D-array. An infinitely indexable ND-array is a data structure that has N dimensions and can store values at any arbitrary combinations of indexes in those dimensions. In practice, an infinitely indexable ND-array can be created using maps or dictionaries. The recent access time of the hash code $b^{(l)}$ can be retrieved by selecting the maximum value in $tstamp^{(l)}[b^{(l)}]$.
%The timestamp when a sample was last hashed into a bucket $b^{(l)}$ is stored in $\_tstamp^{(l)}[b^{(l)}]$, while $tstamp^{(l)}[b^{(l)}][y]$ keeps the timestamp when a sample of class $y$ is hashed in $b^{(l)}$.
The count of samples in bucket $b^{(l)}$ is stored in another infinitely indexable 2D-array $counts^{(l)}$ at an index $counts^{(l)}[b^{(l)}]$. The information required to break the ties during prediction is stored in $sCounts^{(l)}$ and $sAcc^{(l)}$. The variable $sCounts^{(l)}$, an infinitely indexable 2D-array, keeps track of the number of samples from every class $y$ falling into the bucket $b^{(l)}$. $sAcc^{(l)}$, an infinitely indexable 3D-array, stores the class-wise vector sum of all the samples falling into the bucket $b^{(l)}$. The last dimension in this array belongs to the features in the dataset.

Enhash has two phases. In the first one, Enhash predicts the class of a new sample. Assuming that sample $x$ falls into bucket $b^{(l)}$, its distance from all the cluster centers in the bucket is computed (steps \ref{step::mean} and \ref{step::dist}). The variable $cweights$ accumulates the prediction for a sample $x$ via each estimator. This variable is also an infinitely indexable array so that it can accommodate unseen class labels. To get the prediction for sample $x$ from an estimator $l$, a decayed value of count (weighted by distance) is computed (step \ref{step::pred}). The decay factor $2^{-\lambda \Delta t_{1}}$ , depends upon the decay parameter $\lambda$ and the difference of the current time and the last access time of bucket $b^{(l)}$. The decay factor determines whether the previous value in $counts^{(l)}[b^{(l)}]$ is relevant or not, and hence, introducing the forgetting effect in the algorithm. In effect, if the difference in the time is large, then decay is almost zero, and this emulates local replacement in the bucket $b^{(l)}$~\cite{carmona2011gnusmail}. On the other hand, $\lambda$ reduces the effect of samples (possibly, of the same \textit{concept}) hashing into a bucket. The higher is the value of $\lambda$, then more is the rate of decay. Collectively, $\lambda$, and time difference play an important role in drift detection. The log-transformed value of the prediction is added to $cweights$. A pseudo-count of $1$ is added during log-transformation to handle $0$ or near-zero values in prediction. After accumulating predictions from every estimator, a class with maximum weight is designated as the class of sample $x$ (step \ref{step::classassignment}).

In the second phase, the information is updated in the variables to accommodate recent changes in the \textit{concept}. Assume that a recent sample belongs to class $y$. Hence, the value in $tstamp[b^{(l)}][y]$ is set to the current timestamp. Further, to update the effective count in bucket for class $y$ in $counts^{(l)}[b^{(l)}][y]$, the present value is decayed by the difference of the current time and the last seen time of sample from class $y$ and then incremented by $1$ (steps~\ref{step::cUpdate}- \ref{step::countupdate}). This introduces the forgetting phenomenon during updates and also handles spurious changes. For example, if a bucket was accustomed to seeing samples from a particular class and the sudden arrival of a sample from another class that had not been seen by the bucket for a long time, it would not alter the bucket's prediction abruptly. However, after seeing a few samples from the new class, the bucket's confidence will grow gradually towards the recent trends. The value in $counts^{(l)}[b^{(l)}]$ is normalized for numerical stability. Finally, $sCounts^{(l)}[b^{(l)}[y]$ is incremented by 1 and a new sample $x$ is added in $sAcc^{(l)}[b^{(l)}[y]$.

In essence, samples with a similar \textit{concept} are most likely to have the same hash code and hence, share the bucket. For an evolving stream, thus, different buckets are populated. The weight associated with the bucket is gradually incremented when samples of the same \textit{concept} arrive. The contents of a bucket are more relevant when the \textit{concept} reoccurs in the near future than in a faraway future.

Formally, the temporal nature of the posterior distribution of a sample $x$ belonging to a class $y$ is modeled as the Bayes posterior probability $P(y|x) = P(y) P(x|y)/P(x)$. Let $n$ and $n_{y}$ denote the count of total samples and samples of a class $y$, respectively.  In the proposed method, for a given estimator $l$, $P(y) = n_{y}/n$, $P(x|y) = counts^{(l)}[b^{(l)}][y]/n_{y}$, and $P(x) = \sum_{j}(counts^{(l)}[b^{(l)}][j])/n$. \\Thus, $P(y|x) = counts^{(l)}[b^{(l)}][y]/\sum_{j}(counts^{(l)}[b^{(l)}][j])$ and hence, information in $counts^{(l)}[b^{(l)}][y]$ accounts for \textit{concept drift}. For virtual drift, the contents of $b^{(l)}$ may only be supplemented with the additional information from the distribution. For real drift, however, the previous contents of $b^{(l)}$ may be discarded via the decay factor.

\begin{algorithm}
    \setstretch{1.35}
    \caption{Enhash}\label{algo::methodEdit}
    \begin{algorithmic}[1]
        \STATE \textbf{Input:} Data stream $S$
        \STATE \hspace{10mm} $L \gets \text{Number of estimators}$
        \STATE \hspace{10mm} $\textit{bin-width} \gets \text{Width of bucket}$
        \STATE \hspace{10mm} $\lambda \gets \text{Rate of decay}$
        \STATE \textbf{Initialize:} $t \gets 0$
        \STATE \hspace{15mm} For every estimator $l \in \{1,...,L\}$
        \STATE \hspace{20mm} $counts^{(l)} \gets \text{infinitely indexable 2D-array}$
        \STATE \hspace{20mm} $tstamp^{(l)} \gets \text{infinitely indexable 2D-array}$
        %\STATE \hspace{20mm} $\_counts^{(l)} \gets \text{infinitely indexable array}$
        %\STATE \hspace{20mm} $\_tstamp^{(l)} \gets \text{infinitely indexable array}$
        \STATE \hspace{20mm} $sCounts^{(l)} \gets \text{infinitely indexable 2D-array}$
        \STATE \hspace{20mm} $sAcc^{(l)} \gets \text{infinitely indexable 3D-array}$
        \STATE \textbf{Run:}
        \WHILE{$HasNext(S)$}
        \STATE $(x,y) \gets next(S)$
        \STATE $t \gets t + 1$
        \STATE $cweights \gets \text{array initialized with 0s}$
        \FOR{$l \in \{1,...,L\}$}
        %\STATE $b^{(l)} = \lfloor\frac{1}{\text{\textit{bin-width}}}\sum_{j=1}^{j=d}x_j*W^{(l)}_j + bias^{(l)}\rfloor$ \label{step::hashing}
        \STATE $b^{(l)}$ = generate hash code using (\ref{eq::hash}) for estimator $l$\label{step::hashing}
        %\STATE $decay\_value = decay(count, time\_stamp^{(l)}[h^{(l)}])\times table\_group^{(l)}[h^{(l)}]$
        %\STATE $t_{1} = \max_{j}(tstamp^{(l)}[b^{(l)}][j])$
        %\STATE $\Delta t = t - \_tstamp^{(l)}[b^{(l)}]$
        %\STATE $\_decay = 2^{-\lambda \times \Delta t}$
        \STATE $sMean = \frac{sAcc^{(l)}[b^{(l)}]}{sCounts^{(l)}[b^{(l)}]}$\label{step::mean}
        \STATE $dist = \sqrt{\sum((sMean - x)^{2})}$\label{step::dist}
        \STATE $\Delta t_{1} = t - \max_{j}(tstamp^{(l)}[b^{(l)}][j])$
        \STATE $v = 2^{-\lambda \times \Delta t_{1}} \times \frac{counts^{(l)}[b^{(l)}]}{dist}$\label{step::pred}
        \STATE $cweights = cweights + \log(1+v)$ \label{step::predend}
        %\STATE $decay = 2^{-\lambda \times (t - tstamp^{(l)}[b^{(l)}][y])}$ \label{step::updatestart}
        \STATE $\Delta t_{2} = t - tstamp^{(l)}[b^{(l)}][y]$\label{step::updatestart}
        %\STATE $table\_group^{(l)}[h^{(l)}] =  + decay\_value$

        \STATE $counts^{(l)}[b^{(l)}][y] = 1 +  (2^{-\lambda \times \Delta t_{2}} \times counts^{(l)}[b^{l)}][y])$\label{step::cUpdate}
        %\STATE $\_counts^{(l)}[b^{(l)}] = 1 + (\_decay \times \_counts^{(l)}[b^{l)}])$
        \STATE $counts^{(l)}[b^{(l)]} = \frac{counts^{(l)}[b^{(l)}]}{\sum_{j}(counts^{(l)}[b^{(l)}][j])}$\label{step::countupdate}
        %\STATE $time\_stamp^{(l)}[h^{(l)}] = c$
        \STATE $tstamp^{(l)}[b^{(l)}][y] = t$
        %\STATE $\_tstamp^{(l)}[b^{(l)}] = t$
        \STATE $sCounts^{(l)}[b^{(l)}][y] = 1 +  sCounts^{(l)}[b^{(l)}][y]$
        \STATE $sAcc^{(l)}[b^{(l)}][y] = x + sAcc^{(l)}[b^{(l)}][y]$ \label{step::updateend}
        \ENDFOR
        \STATE $\hat{y} \gets \arg\max(cweights)$\label{step::classassignment}
        \ENDWHILE
    \end{algorithmic}
\end{algorithm}

\subsection{Time Complexity Analysis}

There are three important steps in Algorithm~\ref{algo::methodEdit}, namely, the computation of hash function (step~\ref{step::hashing}), prediction of class (steps~\ref{step::mean}-\ref{step::predend}), and update of model parameters (steps~\ref{step::updatestart}-\ref{step::updateend}). In Enhash, the hash function computation is the dot product of a sample with weight parameters (\ref{eq::hash}). Let the time required to compute the hash function be given by $\psi(d)=\mathcal{O}(d)$. The prediction of a sample's class involves the computation of distances from cluster centers in the bucket (\ref{eq::distance}). Assuming that, Enhash has observed $C$ concept classes so far; then, the time required for prediction is given by $\phi(d, C)=\mathcal{O}(d*C)$. The update of the model involves updating the timestamp of the bucket, the effective count of the \textit{concept class} in the bucket, the total samples hashed into the bucket, and accumulation of samples. Effective counts are also normalized in the update step which depends upon $C$. Say, the time required to update model parameter is given by $\zeta(d, C)=\mathcal{O}(d+C)$. Thus, an estimator takes time of order $\mathcal{O}(\psi(d) + \phi(d, C) + \zeta(d, C)) = \mathcal{O}(d+d*C+d+C)=\mathcal{O}(d*C)$. These steps are repeated by all $L$ estimators, and hence, the total time complexity of Enhash, for an arbitrary data set and hyper-parameters settings, is given by $\mathcal{O}(L*d*C)$.

%Arguably, for a given data set of $d$ dimension and for a fixed setting of hyper-parameters, time complexity order of the method reduces to $\mathcal{O}(C)$ with $L*d$ serving as a constant. If number of concept classes are fixed, then processing of newly arrived sample can be performed in $\mathcal{O}(1)$ time.

Thus, for a fixed setting of hyper-parameter $L$, the time complexity of Enhash to process a newly arrived sample for an arbitrary data set is given by $\mathcal{O}{(d*C)}$. However, it can be argued that for a fixed data set, the dimension of data $d$ and number of concept classes $C$ are fixed. Hence, Enhash will effectively take only a constant time, $\mathcal{O}(1)$, in the processing of a new sample.

%In Enhash, say, for an estimator hash function computation for incoming sample takes $\psi(d)$ time, for predicting its class takes $\phi(d)$ time and

\section{Experimental Setup}\label{setup}
Enhash's performance on \textit{concept drift} detection was compared with some of the widely used ensemble learners. These include Learn++~\cite{learn++}, $\text{Learn}^{++}\text{.NSE}$~\cite{learnNse}, Accuracy-Weighted Ensemble (AWE)~\cite{accuracyWeightedEnsemble}, Additive Expert Ensemble (AEE)~\cite{additiveExpertEnsemble}, DWM~\cite{dwm}, Online Bagging-ADWIN (OB)~\cite{adwin,oza2005online}, Leveraging Bagging~\cite{leverageBagging}, Online SMOTE Bagging (OSMOTEB)~\cite{onlineSmotebagging}, and ARF~\cite{arf}. The implementation of these methods is available in \texttt{scikit-multiflow} \texttt{python} package~\cite{skmultiflow}. A fixed value of \textit{estimators} was used for all the methods. For methods such as Accuracy-Weighted Ensemble, $\text{Learn}^{++}\text{.NSE}$, Learn++, Leveraging Bagging, and Online Bagging-ADWIN, the maximum size of window was set $\min(5000, 0.1*n)$, where $n$ is the total number of samples. For the rest of the parameters, the default value was used for all the methods.

\subsection{Evaluation metrics for performance comparison}
We evaluated all the experiments in terms of time, memory consumption, and classifiers' performance. The memory consumption is measured in terms of RAM-hours~\cite{bifet2010fast}. Every GB of RAM employed for an hour defines one RAM-hour. The performance of a classifier is measured in terms of accuracy/error, Kappa M, and Kappa Temporal~\cite{bifet2015efficient}. Kappa M, and Kappa Temporal handle imbalanced data streams, and data streams that have temporal dependencies, respectively. We evaluated the classifiers' performance using the Interleaved Test-Train strategy~\cite{losing2016knn}. This strategy is commonly employed for incremental learning since every sample is used as a test and a training point as it arrives.

\subsection{Description of data sets}
We used 6 artificial/synthetic (Samples x Features)- transientChessboard (200000    x 2), rotatingHyperplane (200000 x 10), mixedDrift (600000 x 2), movingSquares (200000 x 2), interchangingRBF (200000 x 2), interRBF20D (201000 x 20), and 4 real data sets- airlines (539383 x 7), elec2 (45312 x 8), NEweather (18159 x 8), outdoorStream (4000 x 21) for all experiments. The data sets are available on \url{https://github.com/vlosing/driftDatasets}. The synthetic data sets simulate drifts such as abrupt, incremental, and virtual. The real data sets have been used in the literature to benchmark \textit{concept drift} classifiers. The count of samples varies from 4000 (in outdoorStream) to 600,000 (in mixedDrift). Also, outdoorStream has the maximum number of classes, i.e., 40. The summary of the description of data sets is available in Table~\ref{description}.

\subsection{System details}
All experiments were performed on a workstation with 40 cores using Intel Xeon E7-4800 (Haswell-EX/Brickland Platform) CPUs with a clock speed of 1.9 GHz, 1024 GB DDR4-1866/2133 ECC RAM and Ubuntu 14.04.5 LTS operating system with the 4.4.0-38-generic kernel.

\begin{table}
    \centering
        \resizebox{1\textwidth}{!}{
    \begin{tabular}{c c c c}
        \toprule
        Synthetic data sets & Samples x Features & Classes & Drift\\
        \midrule
        transientChessboard & 200,000 x 2 & 8 & Virtual\\
        rotatingHyperplane & 200,000 x 10 & 2 & Abrupt\\
        mixedDrift & 600,000 x 2 & 15 & Incremental, Abrupt, and Virtual\\
        movingSquares & 200,000 x 2 & 4 & Incremental\\
        interchangingRBF & 200,000 x 2 & 15 & Abrupt\\
        interRBF20D & 201,000 x 20 & 15 & Abrupt\\
    \bottomrule
    \end{tabular}
    \hspace{1cm}
    \begin{tabular}{c c c c}
        \toprule
        Real data sets & Samples x Features & Classes\\
        \midrule
        airlines & 539,383 x 7 & 2\\
        elec2 & 45,312 x 8 & 2\\
        NEweather & 18,159 x 8 & 2\\
        outdoorStream & 4,000 x 21 & 40\\
    \bottomrule
    \end{tabular}}
    \caption{Description of data sets.}
    \label{description}
\end{table}

\section{Tuning of parameters for Enhash}\label{sec::tuningEnhash}
The hyperparameters that govern the performance of Enhash constitute $L$ (number of estimators), and $\textit{bin-width}$ (quantization parameter). Even though the decay rate $\lambda$ is also one of the hyperparameters, its value does not require much tuning, and usually, $\lambda=0.015$ is preferred~\cite{sathe2016subspace}. However, making $\lambda=0$ is equivalent to removing the forgetting phenomenon and hence, worsens the performance (discussed further in Section~\ref{sec::ablation}).

\subsection{Constraints to tune L}
\begin{figure}
    %\centering
    \makebox[1 \textwidth][c]{
    \resizebox{1.1 \linewidth}{!}{ %
    \includegraphics[width=\textwidth,keepaspectratio]{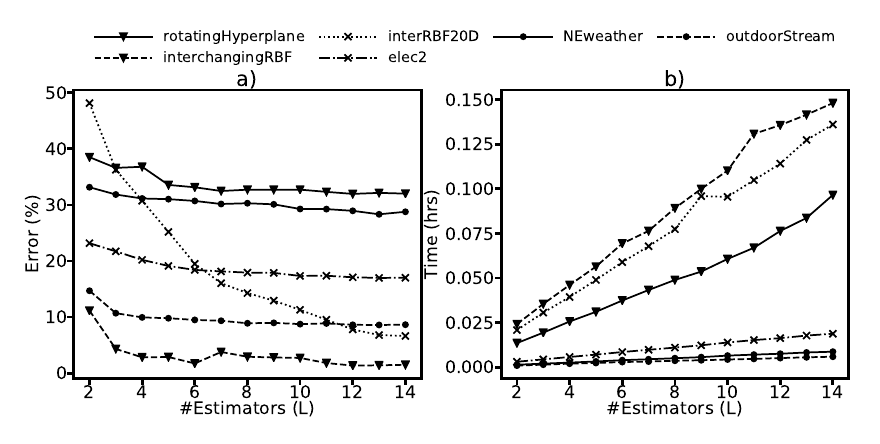}
    }}
    \caption{Tuning of $L$. For artificial, and real data sets, we show a trend in performance metrics, Error (\%), and Time(hrs) for an increase in values of $L$. The value of $L$ varies from [2, 14]. For a given value of $L$, the running time is measured across all the samples for all the estimators in a configuration. The value of error is evaluated across all the samples for a given value of $L$.}
    \label{fig:tuneL}
\end{figure}

In Figure~\ref{fig:tuneL}, it is shown empirically that with an increase in the value of $L$, the performance of Enhash eventually saturates. The time taken by Enhash also increases with $L$. This may be attributed to the fact that for every sample arriving at time $t$, the insertion involves calculating the hash code of the sample for every estimator $l$. It should be emphasized that, as shown empirically in Figure~\ref{fig:tuneL}, only a few estimators are needed to achieve optimum performance. Further, the increase in time due to an increase in $L$ can be reduced through a parallel implementation of Enhash. In that case, evaluation of a hash code for a sample for each $l$ can be done independently of others. Consequently, a moderate value of $L = 10$ is used to perform all the experiments.

\subsection{Constraints to tune \textit{bin-width}}
\begin{figure}
    %\centering
    \makebox[1 \textwidth][c]{
    \resizebox{1.1 \linewidth}{!}{ %
    \includegraphics[width=\textwidth,keepaspectratio]{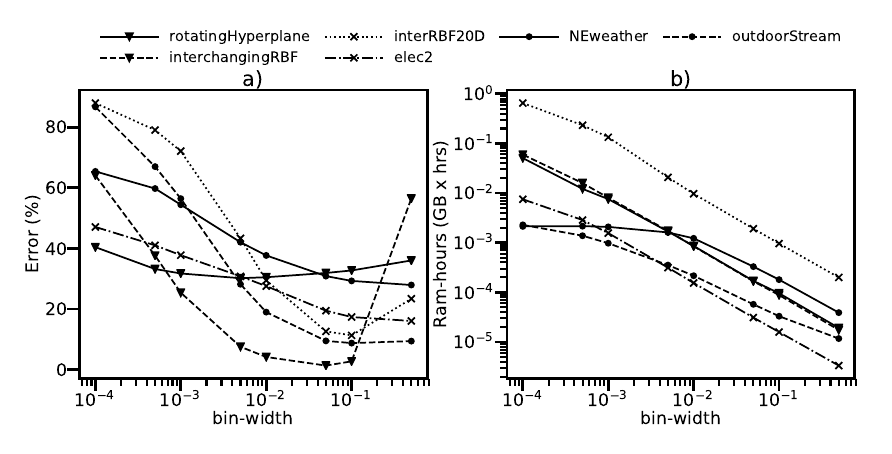}
    }}
    \caption{Tuning of $\textit{bin-width}$. For artificial, and real data sets, we show a trend in performance metrics, Error (\%), and Ram-hours for different values of $\textit{bin-width}$. The value of $\textit{bin-width}$ varies in [0.0001, 0.0005, 0.001, 0.005, 0.01, 0.05, 0.1, 0.5]. a) The value of error is evaluated across all the samples for a given value of $\textit{bin-wdith}$. The values of $\textit{bin-width}$ on x-axis are on logarithmic scale. b) Ram-hours is calculated across all the samples for all the estimators for a given configuration. The values of $\textit{bin-width}$ on x-axis and Ram-hours on y-axis, both are on log scale.}
    \label{fig:tunebinwidth}
\end{figure}

%The value of $\textit{bin-width} \in (0, 1)$
The parameter $\textit{bin-width}$ divides the space into equally spaced unbounded regions of size $\textit{bin-width}$. The smaller is the value of $\textit{bin-width}$, the more granular is the division of space. In other words, for smaller values of $\textit{bin-width}$, the possible values of different hash codes (or buckets) increase rapidly. In the worst case, every sample may fall into a different bucket. Thus, the overall cost to store the contents of all buckets for every estimator grows exponentially. Figure~\ref{fig:tunebinwidth} shows empirically the effect of $\textit{bin-width}$ on memory consumption. The figure highlights that for small values of $\textit{bin-width}$, the overall memory requirement is extremely high. Also, the prediction of the \textit{concept} for the sample may be arbitrary for extremely small values of $\textit{bin-width}$ since there is no neighborhood information in the bucket. As a result, for every new sample, a new \textit{concept} may be falsely predicted. On the other extreme, for large values of $\textit{bin-width}$, samples from different classes may lead to frequent collision. As a result of this, the \textit{concept} of an arriving sample may not be predicted correctly due to confusion in the bucket. Thus, in general, an intermediate range of values for $\textit{bin-width} \in \{0.01, 0.1\}$ is more suitable for all data sets.

\begin{table*}
\makebox[1 \textwidth][c]{
\resizebox{1.1 \linewidth}{!}{ %
    %\centering
    \begin{tabular}{l@{\hspace{1em}} r@{\hspace{1em}} r@{\hspace{1em}} r@{\hspace{1em}} r@{\hspace{1em}} r@{\hspace{1em}} r@{\hspace{1em}} r@{\hspace{1em}} r@{\hspace{1em}} r@{\hspace{1em}} r}
        \toprule
        Dataset & DWM & $\text{Learn}^{++}$ & Learn++ & LB & OB & OSMO- & AWE & AEE & ARF & Enhash \\
        & & .NSE & & & & -TEB & & & & \\
        \midrule
        transientChessboard&49.89&{\textcolor{darkgray}{3.67}}&\textbf{2.75}&13.13&29.62&24.28&89.78&85.37&27.56&18.84\\

        rotatingHyperplane&\textbf{10.11}&24.13&19.36&24.65&{\textcolor{darkgray}{15.08}}&19.75&16.27&18.10&16.40&32.72\\

        mixedDrift&65.77&39.46&49.67&{\textcolor{darkgray}{16.71}}&23.63&20.30&77.79&81.17&19.79&\textbf{12.88}\\

        movingSquares&{\textcolor{darkgray}{29.03}}&68.74&67.84&55.69&65.42&61.12&67.51&67.23&58.54&\textbf{13.29}\\

        interchangingRBF&7.02&75.43&82.79&4.61&37.67&22.05&83.49&82.58&{\textcolor{darkgray}{3.22}}&\textbf{2.72}\\

        interRBF20D&7.62&76.36&81.59&{\textcolor{darkgray}{5.62}}&37.02&21.00&84.90&82.35&\textbf{2.39}&11.30\\

        airlines&{\textcolor{darkgray}{37.42}}&42.92&37.53&43.36&39.23&40.66&38.49&39.17&\textbf{33.09}&41.66\\

        elec2&20.62&34.63&31.40&18.64&23.26&24.60&39.64&26.71&\textbf{11.59}&{\textcolor{darkgray}{17.34}}\\

        NEweather&29.52&29.20&24.63&27.20&\textbf{20.84}&22.74&30.72&30.78&{\textcolor{darkgray}{21.43}}&29.27\\

        outdoorStream&57.62&-&60.88&{\textcolor{darkgray}{9.33}}&34.15&21.25&78.55&42.45&26.12&\textbf{8.75}\\

        \bottomrule
    \end{tabular}
    }
    }
    \caption{Error (in \%) is reported to compare the performance of Enhash with other methods. For a given data set, the method with the least error is in boldface. Due to implementation constraint, $\text{Learn}^{++}\text{.NSE}$ could not run for the outdoorStream data set.}
    \label{errort}
\end{table*}

\section{Experimental Results}\label{results}
For all methods, the number of $estimators$ is considered as 10. In addition for Enhash, \textit{bin-width} was set to \{$0.1$, $0.01$\} and $\lambda$ was set to 0.015. Tables~\ref{errort},~\ref{kappamt}, and~\ref{kappatt} compare the performance of the methods in terms of error, KappaM, and KappaT respectively using Interleaved Test-Train strategy. For these measures, the performance of the proposed method was superior to Accuracy-Weighted Ensemble (AWE), and Additive Expert Ensemble (AEE) on 8 data sets, DWM, $\text{Learn}^{++}\text{.NSE}$, Online SMOTE Bagging (OSMOTEB), and Online Bagging-ADWIN (OB) on 7 data sets, Learn++, and Leveraging Bagging (LB) on 6 data sets, and ARF on 5 data sets. The performance of Enhash supersedes all other methods for 4 data sets - mixedDrift, movingSquares, interchangingRBF, and outdoorStream.

\begin{table*}
\makebox[1 \textwidth][c]{
\resizebox{1.1 \linewidth}{!}{ %
    %\centering
    \begin{tabular}{l@{\hspace{1em}} r@{\hspace{1em}} r@{\hspace{1em}} r@{\hspace{1em}} r@{\hspace{1em}} r@{\hspace{1em}} r@{\hspace{1em}} r@{\hspace{1em}} r@{\hspace{1em}} r@{\hspace{1em}} r}
        \toprule
        Dataset & DWM & $\text{Learn}^{++}$ & Learn++ & LB & OB & OSMO- & AWE & AEE & ARF & Enhash \\
        & & .NSE & & & & -TEB & & & & \\
        \midrule
        transientChessboard&0.43&\textcolor{darkgray}{0.96}&\textbf{0.97}&0.85&0.66&0.72&-0.03&0.02&0.68&0.78\\

        rotatingHyperplane&\textbf{0.80}&0.52&0.61&0.51&\textcolor{darkgray}{0.70}&0.60&0.67&0.64&0.67&0.35\\

        mixedDrift&0.23&0.54&0.42&\textcolor{darkgray}{0.80}&0.72&0.76&0.09&0.05&0.77&\textbf{0.85}\\

        movingSquares&\textcolor{darkgray}{0.61}&0.08&0.10&0.26&0.13&0.19&0.10&0.10&0.22&\textbf{0.84}\\

        interchangingRBF&0.92&0.18&0.10&0.95&0.59&0.76&0.09&0.10&\textcolor{darkgray}{0.96}&\textbf{0.97}\\

        interRBF20D&0.92&0.17&0.11&\textcolor{darkgray}{0.94}&0.60&0.77&0.08&0.10&\textbf{0.97}&0.88\\

        airlines&\textcolor{darkgray}{0.16}&0.04&\textcolor{darkgray}{0.16}&0.03&0.12&0.09&0.14&0.12&\textbf{0.26}&0.06\\

        elec2&0.51&0.18&0.26&0.56&0.45&0.42&0.07&0.37&\textbf{0.73}&\textcolor{darkgray}{0.59}\\

        NEweather&0.06&0.07&0.22&0.13&\textbf{0.34}&0.28&0.02&0.02&\textcolor{darkgray}{0.32}&0.07\\

        outdoorStream&0.41&-&0.37&\textcolor{darkgray}{0.90}&0.65&0.78&0.19&0.56&0.73&\textbf{0.91}\\

        \bottomrule
    \end{tabular}
    }
    }
    \caption{KappaM is tabulated to compare the performances of the methods. For a given data set, the method with the highest value of KappaM is in boldface.}
    \label{kappamt}
\end{table*}

\begin{table*}
\makebox[1 \textwidth][c]{
\resizebox{1.1 \linewidth}{!}{ %
    %\centering
    \begin{tabular}{l@{\hspace{1em}} r@{\hspace{1em}} r@{\hspace{1em}} r@{\hspace{1em}} r@{\hspace{1em}} r@{\hspace{1em}} r@{\hspace{1em}} r@{\hspace{1em}} r@{\hspace{1em}} r@{\hspace{1em}} r}
        \toprule
        Dataset & DWM & $\text{Learn}^{++}$ & Learn++ & LB & OB & OSMO- & AWE & AEE & ARF & Enhash \\
        & & .NSE & & & & -TEB & & & & \\
        \midrule
        transientChessboard&-0.17&\textcolor{darkgray}{0.91}&\textbf{0.94}&0.69&0.30&0.43&-1.11&-1.00&0.35&0.56\\

        rotatingHyperplane&\textbf{0.80}&0.52&0.61&0.50&\textcolor{darkgray}{0.70}&0.60&0.67&0.64&0.67&0.34\\

        mixedDrift&0.28&0.57&0.46&\textcolor{darkgray}{0.82}&0.74&0.78&0.15&0.11&0.78&\textbf{0.86}\\

        movingSquares&\textcolor{darkgray}{0.71}&0.31&0.32&0.44&0.35&0.39&0.32&0.33&0.41&\textbf{0.88}\\

        interchangingRBF&0.92&0.19&0.11&0.95&0.60&0.76&0.11&0.12&\textcolor{darkgray}{0.97}&\textbf{0.97}\\

        interRBF20D&0.92&0.18&0.13&\textcolor{darkgray}{0.94}&0.60&0.78&0.09&0.12&\textbf{0.97}&0.88\\

        airlines&\textcolor{darkgray}{0.11}&-0.02&\textcolor{darkgray}{0.11}&-0.03&0.06&0.03&0.08&0.07&\textbf{0.21}&0.01\\

        elec2&-0.41&-1.36&-1.14&-0.27&-0.59&-0.68&-1.70&-0.82&\textbf{0.21}&\textcolor{darkgray}{-0.18}\\

        NEweather&0.08&0.09&0.23&0.15&\textbf{0.35}&0.29&0.04&0.04&\textcolor{darkgray}{0.33}&0.08\\

        outdoorStream&-4.90&-&-5.23&\textcolor{darkgray}{0.05}&-2.49&-1.17&-7.04&-3.34&-1.67&\textbf{0.10}\\

        \bottomrule
    \end{tabular}
    }
    }
    \caption{KappaT is reported to compare the performances of the methods. The highest value of KappaT in each row is highlighted.}
    \label{kappatt}
\end{table*}

\begin{table*}
\makebox[1 \textwidth][c]{
\resizebox{1.1 \linewidth}{!}{ %
    %\centering
    \begin{tabular}{l@{\hspace{1em}} r@{\hspace{1em}} r@{\hspace{1em}} r@{\hspace{1em}} r@{\hspace{1em}} r@{\hspace{1em}} r@{\hspace{1em}} r@{\hspace{1em}} r@{\hspace{1em}} r@{\hspace{1em}} r}
        \toprule
        Dataset & DWM & $\text{Learn}^{++}$ & Learn++ & LB & OB & OSMO- & AWE & AEE & ARF & Enhash \\
        & & .NSE & & & & -TEB & & & & \\
        \midrule
        transientChessboard&\textcolor{darkgray}{0.169}&0.287&0.477&8.374&0.983&13.623&0.181&0.570&0.481&\textbf{0.099}\\

        rotatingHyperplane&0.205&0.199&0.851&14.022&6.960&114.615&\textcolor{darkgray}{0.198}&0.660&1.318&\textbf{0.067}\\

        mixedDrift&0.757&\textcolor{darkgray}{0.627}&2.726&26.686&7.916&185.500&1.673&17.149&1.875&\textbf{0.339}\\

        movingSquares&\textbf{0.055}&0.179&0.790&9.368&5.936&28.028&0.142&0.567&7.308&{\textcolor{darkgray}{0.068}}\\

        interchangingRBF&\textbf{0.038}&0.185&0.972&8.298&1.444&7.424&0.376&1.048&0.499&{\textcolor{darkgray}{0.132}}\\

        interRBF20D&0.276&\textcolor{darkgray}{0.166}&0.801&19.167&1.507&7.604&2.091&3.200&3.080&\textbf{0.106}\\

        airlines&\textcolor{darkgray}{0.469}&0.571&2.180&37.910&15.151&403.378&0.744&21.669&3.796&\textbf{0.182}\\

        elec2&0.037&0.020&0.160&7.575&1.777&14.511&\textcolor{darkgray}{0.016}&0.059&0.181&\textbf{0.015}\\

        NEweather&0.021&\textcolor{darkgray}{0.008}&0.068&0.590&0.183&1.326&0.013&0.020&0.083&\textbf{0.006}\\

        outdoorStream&0.055&-&\textcolor{darkgray}{0.014}&0.069&0.127&0.162&0.047&0.129&0.071&\textbf{0.004}\\

        \bottomrule
    \end{tabular}
    }
    }
    \caption{The running time of different methods is compared using Time (in hrs). The method with the fastest speed is highlighted for every data set.}
    \label{timet}
\end{table*}

\begin{table*}
\makebox[1 \textwidth][c]{
\resizebox{1.1 \linewidth}{!}{ %
    %\centering
    \begin{tabular}{l@{\hspace{1em}} r@{\hspace{1em}} r@{\hspace{1em}} r@{\hspace{1em}} r@{\hspace{1em}} r@{\hspace{1em}} r@{\hspace{1em}} r@{\hspace{1em}} r@{\hspace{1em}} r@{\hspace{1em}} r}
        \toprule
        Dataset & DWM & $\text{Learn}^{++}$ & Learn++ & LB & OB & OSMO- & AWE & AEE & ARF & Enhash \\
        & & .NSE & & & & -TEB & & & & \\
        \midrule
        transientChessboard&\textbf{1.5e-5}&8.1e-5&\textcolor{darkgray}{3.3e-5}&8.0e-2&3.3e-4&5.5e-1&2.3e-4&5.5e-5&1.2e-3&4.4e-4\\

        rotatingHyperplane&\textbf{3.3e-5}&\textcolor{darkgray}{5.6e-5}&5.7e-5&3.5e-1&1.7e-1&4.6e+1&5.8e-4&1.1e-4&3.4e-2&7.8e-5\\

        mixedDrift&\textbf{1.2e-4}&4.9e-4&\textcolor{darkgray}{3.1e-4}&2.6e-1&2.9e-2&3.9e+1&2.5e-3&2.9e-3&1.3e-2&5.3e-3\\

        movingSquares&\textbf{1.2e-6}&4.9e-5&5.8e-5&9.0e-2&5.5e-2&2.3e+0&1.6e-4&3.3e-5&3.1e+0&\textcolor{darkgray}{1.2e-5}\\

        interchangingRBF&\textbf{7.1e-7}&\textcolor{darkgray}{5.2e-5}&1.1e-4&8.2e-2&7.9e-4&2.4e-1&5.6e-4&1.8e-4&9.3e-4&9.9e-5\\

        interRBF20D&\textbf{3.9e-5}&\textcolor{darkgray}{1.7e-4}&6.4e-4&7.8e-1&2.7e-3&1.1e+0&1.8e-2&4.5e-3&2.6e-3&1.0e-3\\

        airlines&\textbf{5.5e-5}&\textcolor{darkgray}{1.3e-3}&3.4e-3&7.3e-1&2.7e-1&3.0e+2&1.7e-3&2.6e-3&1.5e-1&1.6e-1\\

        elec2&\textcolor{darkgray}{4.9e-6}&\textbf{3.7e-6}&1.1e-5&1.4e-1&3.3e-2&1.3e+0&3.2e-5&8.0e-6&1.5e-3&1.8e-5\\

        NEweather&\textcolor{darkgray}{2.7e-6}&\textbf{7.0e-7}&4.6e-6&4.6e-3&1.4e-3&3.1e-2&1.2e-5&2.6e-6&1.2e-3&1.7e-4\\

        outdoorStream&1.5e-4&-&\textbf{2.0e-6}&2.6e-4&1.9e-3&6.0e-3&1.4e-4&4.5e-4&1.0e-4&\textcolor{darkgray}{3.4e-5}\\

        \bottomrule
    \end{tabular}
    }
    }
    \caption{The memory consumption is measured in terms of RAM-hours. The method with the least value of RAM-hours is highlighted for every data set.}
    \label{ramt}
\end{table*}

Other evaluation criteria are speed (Table~\ref{timet}) and RAM-hours (Table~\ref{ramt}). Table~\ref{timet} reports the overall time (in hrs) taken by each method for a given data set. For a majority of the data sets, Enhash takes the least time. For instance, it took only 0.339 hrs for the mixedDrift data set, followed by $\text{Learn}^{++}\text{.NSE}$, which took 0.627 hrs. Notably, OSMOTEB took more than 185 hrs.

In terms of speed, DWM, $\text{Learn}^{++}\text{.NSE}$, and Learn++ are comparable to Enhash. In terms of accuracy, however, Enhash supersedes the individual methods on the majority of the data sets.  Our findings were consistent across all commonly used evaluation metrics namely error, KappaM, and KappaT. For instance, DWM requires 0.038 hrs on the interchangingRBF data set as compared to 0.132 hrs needed by Enhash. However, the error of Enhash is 2.72\%, while DWM has an error of 7.02\%.

The overall closest competitors of Enhash in terms of evaluation measures error, KappaM, and KappaT are ARF, LB, and OB. Enhash's speed and RAM-hours' requirement are almost insignificant when compared with other methods. For instance, on the movingSquares data set, Enhash needed only 1.2e$-$5 RAM-hours, while ARF, LB, and OB required 3.1, 9.0e$-$2, and 5.5e$-$2 RAM-hours, respectively. Similarly, on the movingSquares data set, Enhash completed the overall processing in 0.068 hrs, while ARF, LB, and OB took 7.308, 9.368, and 5.936 hrs, respectively.

OSMOTE is the slowest when compared with all other methods. For the airlines data set, OSMOTE took more than 403 hrs, while Enhash required only the minimum amount of time of 0.182 hrs.

The remaining methods, AWE, and AEE have inadequate performances as compared to Enhash. On the transientChessboard data set, the error values are as high as 89.78\% and 85.37\% for AWE and AEE, respectively.

In summary, DWM has relatively poor performance in detecting virtual drifts but fairs well in abrupt and incremental drifts. Learn++ is exceptionally well in detecting virtual drift but severely under-performs in abrupt and incremental drifts. LB and ARF suffer in detecting incremental drifts. However, LB and ARF have an overall satisfactory performance. Enhash performs relatively well on all data sets. Enhash has a superior performance on a data set consisting of three different kinds of drifts, namely incremental, virtual, and abrupt drifts. Although Enhash falls short in detecting abrupt drifts, but the performance gap is not very significant.

\section{Ablation study}\label{sec::ablation}

Enhash is built by modifying (\ref{eq::class}) and (\ref{eq::update}). The two major changes in Enhash from these are the inclusion of a forgetting factor and a heuristic for tie braking mechanism to reduce the false positives. In this section, we assess the impact of these changes, which make Enhash suitable for the \textit{concept drift} detection.

Table~\ref{ablation} compares the performances of Enhash with its two variants - Enhash-lambda0 and Enhash-noWeights. Enhash-lambda0 refers to the variant when $\lambda=0$ or equivalently, the forgetting phenomenon is not accounted. Enhash-noWeights refers to the variant of Enhash when ties in the assignment of \textit{concept class} are broken randomly. In other words, the distance (\ref{eq::distance}) of an incoming sample from the mean of the classes in the bucket is not used to determine the class in case of ties. The values of the other hyperparameters are the same as that in Section~\ref{results}. Table~\ref{ablation} shows that the performance of Enhash is much superior to its both variants for the majority of the data sets. The sub-optimal results of Enhash-noWeights may be attributed to the fact that when the same count of samples from different classes is present in the same bucket, the class for an incoming sample gets assigned randomly.

\begin{table*}
\centering
\begin{tabular}{lrrrr}
\toprule
Dataset & Enhash & Enhash-lambda0 & Enhash-noWeights \\
\midrule
transientChessboard&\textbf{18.84}&19.57&30.58\\

rotatingHyperplane&\textbf{30.49}&32.26&36.69\\

mixedDrift&\textbf{12.88}&13.13&16.45\\

movingSquares&11.76&11.68&\textbf{11.66}\\

interchangingRBF&\textbf{2.72}&2.82&3.58\\

interRBF20D&\textbf{11.30}&11.75&12.98\\

airlines&41.66&\textbf{41.36}&43.03\\

elec2&17.34&\textbf{17.28}&17.65\\

NEweather&\textbf{29.27}&30.17&32.19\\

outdoorStream&\textbf{8.75}&\textbf{8.75}&9.55\\

\bottomrule
\end{tabular}
\caption{Ablation study of Enhash. The performance of Enhash is compared with its two different variants- 1. Enhash with $\lambda=0$ (referred to as Enhash-lambda0), and 2. Enhash when ties in \textit{concept class} assignments are not broken by considering the distance of an incoming sample from the mean of classes in the bucket (referred to as Enhash-noWeights).}
\label{ablation}
\end{table*}

\section{Conclusions}\label{conclusion}
We conclude that Enhash supersedes other methods in terms of speed since the algorithm effectively requires only $\mathcal{O}(1)$ running time for each sample on a given estimator. In addition, the performance of Enhash in terms of error, KappaM, and KappaT is better or comparable to others for majority data sets. These data sets constitute abrupt, gradual, virtual, and reoccurring drift phenomena. The closest competitor of Enhash in terms of performance is the Adaptive Random Forest. Notably, Enhash requires, on an average 10 times lesser RAM-hours than that of Adaptive Random Forest.

%\begin{acknowledgements}
%If you'd like to thank anyone, place your comments here
%and remove the percent signs.
%\end{acknowledgements}

% Authors must disclose all relationships or interests that
% could have direct or potential influence or impart bias on
% the work:
%
% \section*{Conflict of interest}
%
% The authors declare that they have no conflict of interest.
\section*{Declarations}

\subsection*{Funding}
Not applicable

\subsection*{Competing interests}
The authors declare that they have no known competing interests.

\subsection*{Availability of data and material}
The data sets are available on  \url{https://github.com/vlosing/driftDatasets}.

\subsection*{Code availability}
Enhash implementation is available at \url{https://www.dropbox.com/sh/7su1s0sfhfb5fhb/AAAneOBtOCi58If6k-WFpIjVa?dl=0}. All experiments in the manuscript have been performed using \texttt{python 3.0}

\subsection*{Author's contributions}
\textbf{Aashi Jindal}: Design of this study, Methodology, Software. \textbf{Prashant Gupta}: Design of this study, Methodology, Software, Investigation. \textbf{Jayadeva}: Conceptualization of this study, Design. \textbf{Debarka Sengupta}: Conceptualization of this study, Design.

% BibTeX users please use one of
%\bibliographystyle{spbasic}      % basic style, author-year citations
\bibliographystyle{spmpsci}      % mathematics and physical sciences
\bibliography{main}   % name your BibTeX data base

\end{document}